\pdfoutput=1
\PassOptionsToPackage{table}{xcolor}
\documentclass[11pt]{article}
\usepackage[final]{ACL2023}
\usepackage{times}
\usepackage{latexsym}
\usepackage[T1]{fontenc}
\usepackage[utf8]{inputenc}
\usepackage{microtype}
\usepackage{inconsolata}
\usepackage{graphicx}
\usepackage[table]{xcolor}
\usepackage{comment}
\definecolor{blue}{HTML}{AADDFF}
\definecolor{cyan}{HTML}{00FFDD}
\definecolor{green}{HTML}{AAFFAA}
\definecolor{yellow}{HTML}{FFFF88}
\definecolor{red}{HTML}{FF8888}
\definecolor{orange}{HTML}{FFBB55}
\definecolor{purple}{HTML}{AAAAFF}
\definecolor{purple2}{HTML}{CDC9FF}
\definecolor{pink}{HTML}{FFAAFF}
\definecolor{grey}{HTML}{CCCCCC}

\title{Robust and Fine-Grained Detection of AI Generated Texts}
\author{
 \textbf{Ram Mohan Rao Kadiyala               \textsuperscript{1,3}},
 \textbf{Siddartha Pullakhandam               \textsuperscript{2}},
 \textbf{Kanwal Mehreen                       \textsuperscript{3}},\\
 \textbf{Siddhant Gupta                       \textsuperscript{3,5}},
 \textbf{Jebish Purbey                        \textsuperscript{3}},
 \textbf{Drishti Sharma                       \textsuperscript{3}},
 \textbf{Ashay Srivastava                     \textsuperscript{4,6}},\\
 \textbf{Subhasya TippaReddy                  \textsuperscript{7}},
 \textbf{Arvind Reddy Bobbili                 \textsuperscript{8}},
 \textbf{Suraj Telugara Chandrashekhar        \textsuperscript{4}},\\
 \textbf{Ankan Ghosh                          },
 \textbf{Modabbir Adeeb                       \textsuperscript{4}},
 \textbf{Srinadh Vura                         \textsuperscript{9}},
 \textbf{Suman Debnath                        \textsuperscript{11}},
 \textbf{Hamza Farooq                         \textsuperscript{1,10}}\\
 \\
 \textsuperscript{1}Traversaal.ai   
 \textsuperscript{2}Vantager  
 \textsuperscript{3}Cohere for AI Community
 \textsuperscript{4}University of Maryland\\
 \textsuperscript{5}IIT Roorkee
 \textsuperscript{6}Perplexity
 \textsuperscript{7}University of South Florida
 \textsuperscript{8}University of Houston\\
 \textsuperscript{9}IISc Bangalore
 \textsuperscript{10}Stanford University
 \textsuperscript{11}Amazon\\
 \\
 \small{\textbf{Correspondence : } \href{mailto:contact@rkadiyala.com}{contact@rkadiyala.com}}\\
 \small{\textbf{Resources : } \href{https://huggingface.co/collections/1024m/text-authorship-attribution-co-authored-texts-66907693cbcaf7ab0e99c413}{Datasets \& Models}}\\
 \small{\textbf{Demo : } \href{https://huggingface.co/spaces/M2ai/MGTD-Demo}{Demo HF Space}}\\
}
\begin{document}
\maketitle
\begin{abstract}
An ideal detection system for machine-generated content is supposed to work well on any generator as many more advanced LLMs come into existence day by day. Existing systems often struggle with accurately identifying AI-generated content over shorter texts. Further, not all texts might be entirely authored by a human or LLM; hence, we focused more on partial cases, i.e., human-LLM co-authored texts. Our paper introduces a set of models built for the task of token classification that are trained on an extensive collection of human-machine co-authored texts, which performed well over texts of unseen domains, unseen generators, texts by non-native speakers, and those with adversarial inputs. We also introduce a new dataset of over 2.4M such texts, mostly co-authored by several popular proprietary LLMs in over 23 languages. We also present findings of our models' performance over texts of each domain and generator. Additional findings include a comparison of performance against each adversarial method, the length of input texts, and the characteristics of generated texts compared to the original human-authored texts.
\end{abstract}
\section{Introduction}
\label{sec:introduction}
Recent advancements in large language models (LLMs) have significantly narrowed the gap between machine-generated and human-authored text. As LLMs continue to improve in fluency and coherence, the challenge of reliably detecting AI-generated content could become increasingly critical. This issue is particularly pressing in domains such as education and online media, where the authenticity of textual material is paramount. While early efforts such as the GLTR \citep{gehrmann-etal-2019-gltr} provided valuable insights by leveraging statistical methods to differentiate between human and machine text, these methods often lag behind the rapid pace of LLM evolution. Likewise, initiatives aimed at mitigating neural fake news \citep {zellers2019defending} have made significant strides in addressing the societal implications of AI-generated misinformation. However, as LLMs become more sophisticated, existing detection systems must be re-evaluated and enhanced to maintain their effectiveness. Further, each domain comes with its version of the issue of detecting machine-generated texts. For instance, proprietary LLMs with internet access and better knowledge cutoffs are more likely to be used in domains like academia. Similarly, bad actors might use an open-source generator for the task of creating misinformation and deception through machine-generated online content, as such models can be hosted locally to not leave a trail and are more flexible in terms of not denying user requests. Hence, tailoring models and approaches for each specific domain/scenario might be more applicable for practical scenarios. We chose a token-classification approach to train a model for the task of distinguishing writing styles within a text if more than one was found. This approach helped us achieve better performance over texts of unseen features (i.e., domain, generator, adversarial inputs, and non-native speakers' texts) as our models were trained to distinguish different styles within a text rather than classifying an input text as one of the two classes it was trained on. Further, we explored the findings and results upon testing our models over other benchmarks, which consist of texts from unseen domains and generators. We also tested our models over benchmarks, which consist of texts with various adversarial inputs and those written by non-native speakers. 
\begin{table*}
    \centering
    \begin{tabular}{ccccc}
    \noalign{\hrule height 2pt}
    \rowcolor{cyan} \textbf{Source}     & \textbf{Dataset/Benchmark} & \textbf{Samples} & \textbf{Languages} & \textbf{Generators} \\
    \noalign{\hrule height 2pt}     
    \citep{lee2022coauthor}     & CoAuthor                 & 1,445        & 1          & 1            \\
    \citep{zhang2024llm}        & MixSet                   & 3,600        & 1          & 12           \\
    \citep{dugan2022realfaketextinvestigating} & RoFT      & 21,646       & 1          & 5            \\
    \citep{macko2024authorship} & MultiTude                & 4,070        & 11         & 8            \\
    \citep{wang-etal-2024-semeval-2024} & M4GTD            & 31,893       & 1          & 4            \\
    \citep{artemova2025beemobenchmarkexperteditedmachinegenerated} & Beemo & 19,600  & 1  & 10        \\
    \noalign{\hrule height 2pt}     
     Our Work                   & \textit{placeholder}    & 2,447,221    & 23         & 12            \\
    \noalign{\hrule height 2pt}     
    \end{tabular}
    \caption{Comparison with other Human-LLM co-authored datasets \& benchmarks}
    \label{table:0}
\end{table*}
\section{Related Works}
\label{sec:relatedworks}
A major portion of current research in detecting machine-generated content focuses on longer-form writing through binary classification. However, AI-generated misinformation is more likely to cause harm than its use in academia, making the distinction between AI and human-generated texts on social media platforms a critical challenge. Existing methods often struggle with accurately identifying AI-generated content over shorter texts. Moreover, binary classification approaches, which categorize texts as either human or AI-generated, \citep{wang2024semeval2024task8multidomain, wang2024m4gtbenchevaluationbenchmarkblackbox, bhattacharjee2023condacontrastivedomainadaptation, NEURIPS2019_3e9f0fc9, Macko_2023, ghosal2023towards, dugan2024raid} are less practical in settings where texts could be co-authored by both humans and LLMs. In contrast, binary classification may be more effective for shorter texts commonly found on reviews and social media platforms \citep{macko2024multisocialmultilingualbenchmarkmachinegenerated, ignat2024maideupmultilingualdeceptiondetection}, where content typically consists of one or two sentences. Additionally, some detection works rely on detecting watermarks from AI-generated texts. \citep{chang2024postmark, dathathri2024scalable, sadasivan2024aigeneratedtextreliablydetected, zhao2023provable} but not all generators utilize watermarking, limiting the applicability of such approaches. A few other approaches utilize statistical methods \citep{mitchell2023detectgpt, kumarage2023stylometric, gehrmann2019gltr, hans2024spottingllmsbinocularszeroshot, bao2023fast}, but they can be prone to misclassification against adversarial methods like rephrasing and humanizing. \citep{abassy2024llmdetectaivetoolfinegrainedmachinegenerated} introduced a 4-way classification as entirely human-authored, entirely LLM-authored, human-edited and LLM-authored, or LLM-edited human-authored. An ideal detection system should be capable of identifying AI-generated content from any generator without depending on watermarking, especially since watermarking techniques may not be effective for shorter texts. Further, an ideal detector should be robust against adversarial methods. To properly deal with co-authored text cases, a token classification approach to detect boundaries \citep{dugan2022realfaketextinvestigating}, \citep{macko2024authorship} between machine-authored and human-authored portions might be more appropriate. Further, in cases of AI usage in scenarios like academic cases, users are likely to use a proprietary LLM with better knowledge cutoffs than an open-source LLM. Similarly, for AI misuse over social platforms, users are more likely to use an open-source model due to better flexibility and privacy. Hence, building models and benchmarks with an appropriate set of LLMs might be more applicable for practical scenarios. Many proprietary systems struggle at the task of fine-grained detection; further, a large enough dataset to cover all POS-tag bi-grams of the text boundaries is required for such fine-grained detectors to work well \citep{kadiyala-2024-rkadiyala}. Previous works in the similar direction include \citep{lee2022coauthor, zhang2024llm, Dugan2023realorfake, macko2024authorship, liang2024monitoring, chen2023gpt}, which utilize a dataset of limited size and limited number of generators or those less likely to be used, which might not be enough for a detector to work well on unseen domains and generators' texts. Further, the task of detection of such human-LLM co-authored texts is a harder task compared to binary classification of texts based on authorship \citep{geng2025humanllmcoevolutionevidenceacademic, 10.1145/3715073.3715076}. 
\section{Dataset}
\label{sec:dataset}
Our dataset consists of around 2.45 M samples. We used 12 different LLMs, out of which 9 are popular proprietary LLMs: GPT-o1 \citep{openai_system_card}, GPT-4o \citep{openai2024gpt4technicalreport}, Gemini-1.5-Pro \citep{deepmind2023gemini}, Gemini-1.5-Flash, Claude-3.5-Sonnet \citep{anthropic2023claude3}, Claude-3.5-Haiku, Perplexity-Sonar-Large \citep{perplexity2023sonar}, Amazon-Nova-Pro \citep{Intelligence2024}, and Amazon-Nova-Lite. We also included 3 open-source LLMs, i.e., Aya-23 \citep{aryabumi2024aya23openweight}, Command-R-Plus \citep{cohere_for_ai_2024}, and Mistral-large-2411 \citep{mistral2024mistrallarge}, which produced outputs that are relatively difficult to distinguish from human-written texts compared to other similar models in other benchmarks\footnote{\url{https://raid-bench.xyz/}} as well as our own datasets. The samples range from 30 to 25,000 words in length with an average length of around 600 words. \autoref{table:0} provides an overview of comparison with other datasets and benchmarks for fine-grained detection of human-LLM co-authored texts. Our dataset utilizes a better choice of generators, which are more likely to be used in practical scenarios, whereas prior datasets are limited to smaller and a limited number of models. Our dataset also comprises the largest collection of human-LLM co-authored texts, aiding other researchers in the field.
\subsection{Dataset Distribution}
\label{subsec:datasetdistribution}
\begin{figure*}[!h]
    \centering
    \includegraphics[width=0.8\linewidth]{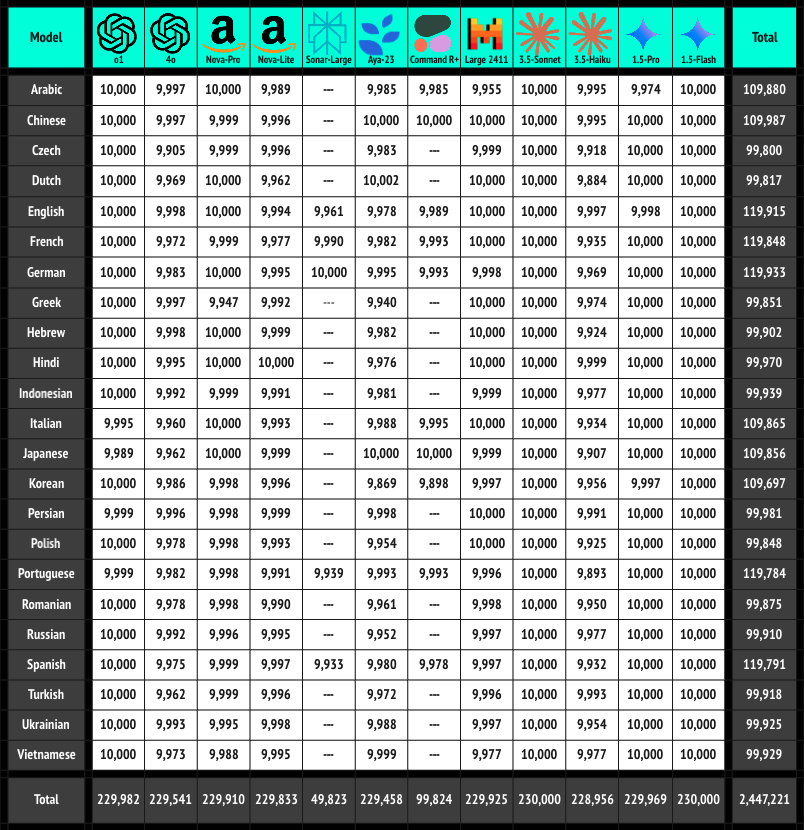}
    \caption{Dataset distribution per each generator and language in our dataset}
    \label{figure:1}
\end{figure*}

The language distribution of the dataset and LLMs used can be seen in \autoref{figure:1}. Each language-LLM pair has roughly 10000 samples. Among each set of the 10000 samples, training, development, and test sets constitute 40\%, 10\%, and 50\%, respectively. Additionally, among each set of 10000 samples, 10\% were completely human written, another 10\% completely machine generated, and the other 80\% were human-LLM co-authored, i.e., a few portions of the text are machine generated and the rest are human written. 
\subsection{Dataset Creation}
\label{subsec:datasetcreation}
GPT-4o was used through an Azure OpenAI endpoint\footnote{\url{https://azure.microsoft.com/en-us/products/ai-services/openai-service/}}. Command-r-plus and Aya-23 were used through Cohere's API platform\footnote{\url{https://dashboard.cohere.com/}}. The rest of the models were used through OpenRouter's\footnote{\url{https://openrouter.ai/models}} API. The rewritten samples were created by providing the generator LLM with the original text and a random prompt among writing an alternate version, a later update of what happened, or a rephrased version of the same text. The samples that returned the exact text or a very similar text were once again regenerated. The partially machine-generated texts were created by splitting the text at random locations, and the generator was asked to finish the text. The split locations were chosen randomly from between the 30th word up to the end of the text. This was done to provide the LLM with enough context to better work towards text completion.
\subsection{Original Data Source and Filtering}
\label{subsec:originaldatasourcesandfiltering}
With a goal of training on one domain and testing on every other, we chose to train on old newspapers \citep{alvations_oldnewspapers} as it has a sufficient number of samples, i.e., 17.2 M for 67 languages of the same domain. We then removed samples that originated after the release of GPT-3 to avoid mislabeling of samples in our dataset. Further, we sampled texts that were at least 3 sentences or 50 words long. For Chinese and Japanese, we sampled texts that were at least 100 characters long.
\section{Our System}
\label{sec:oursystem}
We have experimented with various multilingual transformer models \citep{he2023debertav}, \citep{conneau2019unsupervised}, \citep{beltagy2020longformer} with/without additional LSTM \citep{hochreiter1997long} or CRF layers \citep{zheng2015conditional} through a binary token-classification approach. We found that using an additional CRF layer produced better results compared to other setups with the same model. All of the transformer models tested have produced nearly identical results over our test set. However, XLM-Longformer gave better results over unseen domains and generators' texts and was used in the end given the longer default context length of 16384. The token-level predictions by the models were then mapped into word-level predictions. We use the model's predictions to separate text portions based on perceived authorship. Improving the performance required balancing pre- and post-boundary POS tag bi-grams to reduce error rates \citep{kadiyala-2024-rkadiyala}.
\section{Evaluation and Results}
\label{subsec:evaluation}
We evaluate the models at 3 levels of granularity: word level, sentence level, and overall. For Chinese and Japanese, we performed evaluation at a character level instead of a word level. \footnote{Unlike English, Chinese and Japanese are written without spaces between words, so segmenting text into words requires an external lexicon or trained segmenter and can introduce alignment errors. Using characters as the atomic units avoids these segmentation ambiguities and ensures every unit is evaluated consistently}. Each domain and user might have a different preference towards metrics and evaluation; hence, we report 3 metrics at each level of granularity: accuracy, recall, and precision. For word-level mapping of predictions, in cases where part of a word, i.e., a few tokens, are classified differently than others, we assign the same label to the word as its first token. While mapping word-level predictions to a sentence, we used majority voting, and in cases where consensus was not obtained, we assigned the same label as the first word. For evaluation over other benchmarks requiring binary classification of texts as human or machine written, we assign a human-written label to the text if at least one percent of the words get classified as human-written. We also report several metrics, some of which can be seen in the below tables; the rest can be found in \autoref{sec:otherresults}. 
\subsection{Seen Domains \& Seen Generators}
\label{subsec:seendomainsandseengenerators}
The results of our models over our dataset's test set can be seen in \autoref{table:1}. The samples from both the data splits are of the same domain and originate from the same set of generators.
\begin{table*}[!h]
    \centering
    \begin{tabular}{ccccc}
    \noalign{\hrule height 2pt}
    \rowcolor{cyan}\textbf{\small{Language }}$\downarrow$ & \textbf{\small{Partial cases}} & \textbf{\small{Unchanged cases}} & \textbf{\small{Rewritten cases}} & \textbf{\small{Overall}} \\
    \noalign{\hrule height 2pt}
    \small{Arabic}            & \small{97.16} & \small{90.69} & \small{97.55} & \small{96.44} \\
    \small{Chinese*}          & \small{93.13} & \small{76.28} & \small{91.40} & \small{86.58} \\
    \small{Czech}             & \small{96.23} & \small{79.63} & \small{93.84} & \small{94.98} \\
    \small{Dutch}             & \small{96.83} & \small{77.60} & \small{94.13} & \small{95.31} \\
    \small{English}           & \small{97.32} & \small{90.23} & \small{97.68} & \small{96.02} \\
    \small{French}            & \small{96.89} & \small{74.46} & \small{96.52} & \small{94.91} \\
    \small{German}            & \small{96.64} & \small{76.54} & \small{95.92} & \small{95.28} \\
    \small{Greek}             & \small{96.25} & \small{82.21} & \small{92.08} & \small{94.37} \\
    \small{Hebrew}            & \small{96.52} & \small{80.56} & \small{95.34} & \small{95.70} \\
    \small{Hindi}             & \small{97.08} & \small{92.60} & \small{97.24} & \small{96.34} \\
    \small{Indonesian}        & \small{97.20} & \small{84.92} & \small{97.19} & \small{96.64} \\
    \small{Italian}           & \small{96.44} & \small{80.69} & \small{96.84} & \small{95.38} \\
    \small{Japanese*}         & \small{92.74} & \small{83.80} & \small{92.81} & \small{86.13} \\
    \small{Korean}            & \small{97.29} & \small{84.13} & \small{94.74} & \small{95.77} \\
    \small{Persian}           & \small{96.60} & \small{88.61} & \small{96.19} & \small{94.36} \\
    \small{Polish}            & \small{96.63} & \small{88.52} & \small{92.75} & \small{95.94} \\
    \small{Portuguese}        & \small{96.46} & \small{88.51} & \small{90.29} & \small{94.89} \\
    \small{Romanian}          & \small{97.59} & \small{78.06} & \small{95.15} & \small{96.10} \\
    \small{Russian}           & \small{96.64} & \small{79.98} & \small{95.58} & \small{94.02} \\
    \small{Spanish}           & \small{96.38} & \small{71.60} & \small{96.69} & \small{94.47} \\
    \small{Turkish}           & \small{95.74} & \small{83.00} & \small{94.48} & \small{93.62} \\
    \small{Ukrainian}         & \small{95.74} & \small{74.03} & \small{96.57} & \small{93.53} \\
    \small{Vietnamese}        & \small{94.41} & \small{77.99} & \small{96.65} & \small{89.67} \\
    \noalign{\hrule height 2pt}
    \rowcolor{grey!25} \textbf{\small{Average}}  & \textbf{\small{96.26}} & \textbf{\small{81.94}} & \textbf{\small{95.11}} & \textbf{\small{94.19}} \\
    \noalign{\hrule height 2pt}     
    \end{tabular}
    \caption{Word-Level Accuracy (.2f) of the models on the test dataset for each case}
    \caption*{* Character level evaluations were done instead for Japanese and Chinese}
    \label{table:1}
\end{table*}
\subsection{Unseen Domains \& Unseen Generators}
\label{subsec:unseendomainsandunseengenerators}
\begin{table}
    \centering
    \begin{tabular}{ccccc}
    \noalign{\hrule height 2pt}
   \rowcolor{cyan} \tiny{\textbf{Metrics }}$\rightarrow$ & \small{\textbf{Accuracy}} & \small{\textbf{Precision}} & \small{\textbf{Recall}} & \small{\textbf{F1}}\\
    \noalign{\hrule height 2pt}
    \tiny{Initial Model}     & \small{86.51} & \small{91.61} & \small{87.46} & \small{89.49} \\
    \tiny{Final Model}       & \small{86.00} & \small{87.16} & \small{92.25} & \small{89.63} \\
    \noalign{\hrule height 2pt}
    \end{tabular}
    \caption{Word level Metrics over Mgtd-bench (.2f) through our models (zero-shot, unseen domains, unseen generators)}
    \label{table:2}
\end{table}
\begin{table}
    \centering
    \begin{tabular}{ccccc}
    \noalign{\hrule height 2pt}
    \rowcolor{cyan}\small{\textbf{Metrics }}$\rightarrow$ & \small{\textbf{Accuracy}} & \small{\textbf{Precision}} & \small{\textbf{Recall}} & \small{\textbf{F1}}\\
    \noalign{\hrule height 2pt}
    \small{Arabic}       & \small{95.9} & \small{96.1} & \small{94.5} & \small{95.2} \\
    \small{English}      & \small{99.1} & \small{98.7} & \small{99.3} & \small{99.0} \\
    \hline 
    \tiny{Arabic-Best}       & \small{96.1} & \small{96.1} & \small{95.0} & \small{95.5} \\
    \tiny{English-Best}      & \small{99.3} & \small{99.0} & \small{99.2} & \small{99.1} \\    
    \noalign{\hrule height 2pt}
    \end{tabular}
    \caption{Overall Metrics over ETS essays (.1f) through our detectors (zero-shot, unseen generators, unseen domain) VS best submissions (fine-tuned on same generators and domain)}
    \label{table:3}
\end{table}
The models were tested twice over \citep{wang2024semeval2024task8multidomain}: once by training on just 10000 samples of a single generator (Aya-23) and again later by training over our complete training data. The benchmark consists of 11,123 samples of peer reviews and student essays \citep{koike2024outfox}; the generators used were various versions of Llama-2 and ChatGPT (an earlier version of GPT-4). The samples would hence be from completely unseen domains and generators to our models. The results of both models can be seen in \autoref{table:2}. 
\subsection{Unseen Domains \& Unseen Generators \& Non-Native Speakers}
\label{subsec:etsessayssharedtask}
The models were tested by training on just 10,000 samples each from Aya-23 for English and Arabic separately. The benchmark's samples for Arabic were from \citep{alfaifi2013arabic} and \citep{zaghouani2024qcaw}. The samples for English consist of ETS and IELTS student essays sampled from non-native speakers \citep{chowdhury-etal-2025-genai}. Our models were used for inference directly over these texts, and the strings of predicted tokens were then used for binary classification based on how frequently the perceived authorship changed from human to LLM and vice versa, i.e., the number of changes and whether the longest string consists of ones or zeroes. The metrics obtained for each language can be seen in \autoref{table:3}. 
\begin{table*}[!h]
    \centering
    \begin{tabular}{cccccc}
    \noalign{\hrule height 2pt}
    \rowcolor{cyan}\textbf{{System}} & \textbf{Source} & \textbf{{M}} & \textbf{{E}} & \textbf{{R}} & \textbf{{Avg}}\\
    \noalign{\hrule height 2pt}     
    {GLTR} & \citep{gehrmann2019gltr}                                & 0.706          & 0.943            & 0.773          & 0.807 \\
    {Binoculars} & \citep{hans2024spottingllmsbinocularszeroshot}    & \textbf{0.822} & \textbf{0.995}   & 0.613          & 0.778 \\
    {RADAR} & \citep{gu2025radarbenchmarkinglanguagemodels}          & 0.768          & 0.939            & 0.716          & 0.794 \\
    \noalign{\hrule height 2pt} 
     \textbf{Current work} & Our Baselines                           & 0.713          & 0.914            & \textbf{0.811} & \textbf{0.812} \\
    \noalign{\hrule height 2pt} 
    \end{tabular}
    \caption{Comparisons with other systems on the same benchmarks: MGTD-Bench (\textbf{M}), Academic-Essays (\textbf{E}), and Raid-Bench (\textbf{R}) through F1 Scores, and the average scores from the 3 benchmarks (\textbf{AVG}). The results cover a randomly sampled subset of each benchmark spanning over 13k samples, including those with adversarial methods (\textbf{R}) and non-native speakers (\textbf{E}). Randomly sampled subsets were used with equal distribution of all generators, domains, and adversarial methods distribution in the subset for each benchmark.}
    \label{table:1000}
\end{table*}
\subsection{Unseen Domains \& Partially Seen Generators \& Adversarial Inputs}
\label{subsec:raidbenchsharedtask}
\begin{figure}[!h]
    \centering
    \includegraphics[width=0.95\linewidth]{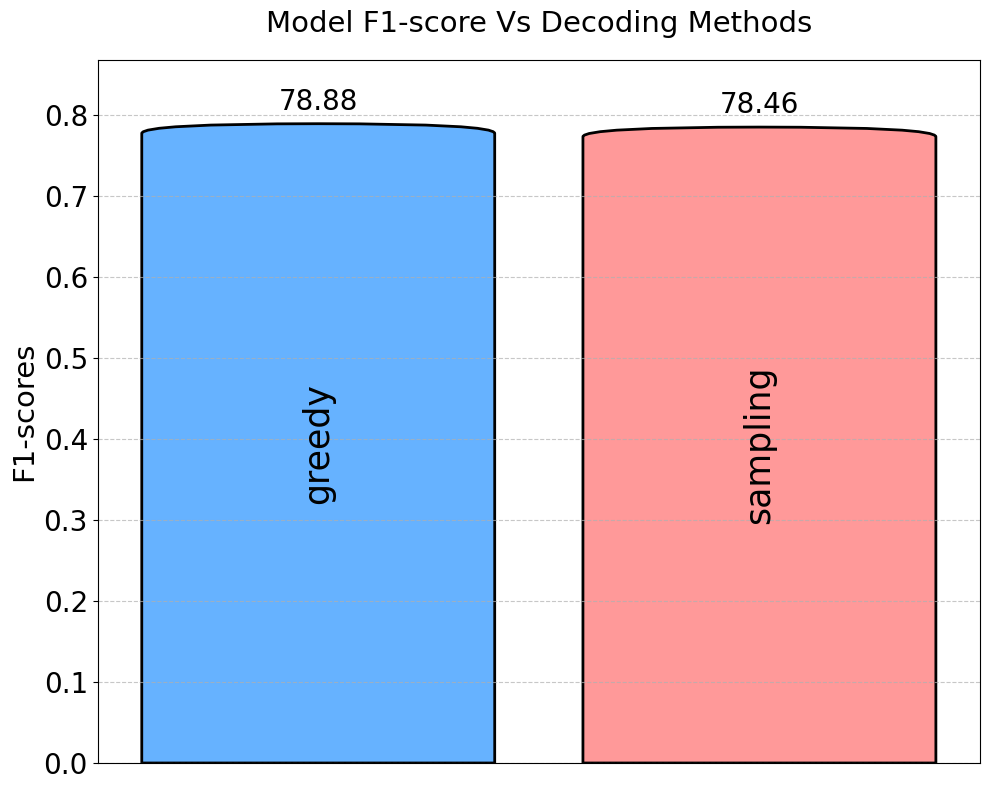}
    \caption{F1 scores VS text sampling method used : Sampling strategy did not effect detection capability}
    \label{figure:01}
\end{figure}

\begin{figure}[!h]
    \centering
    \includegraphics[width=1\linewidth]{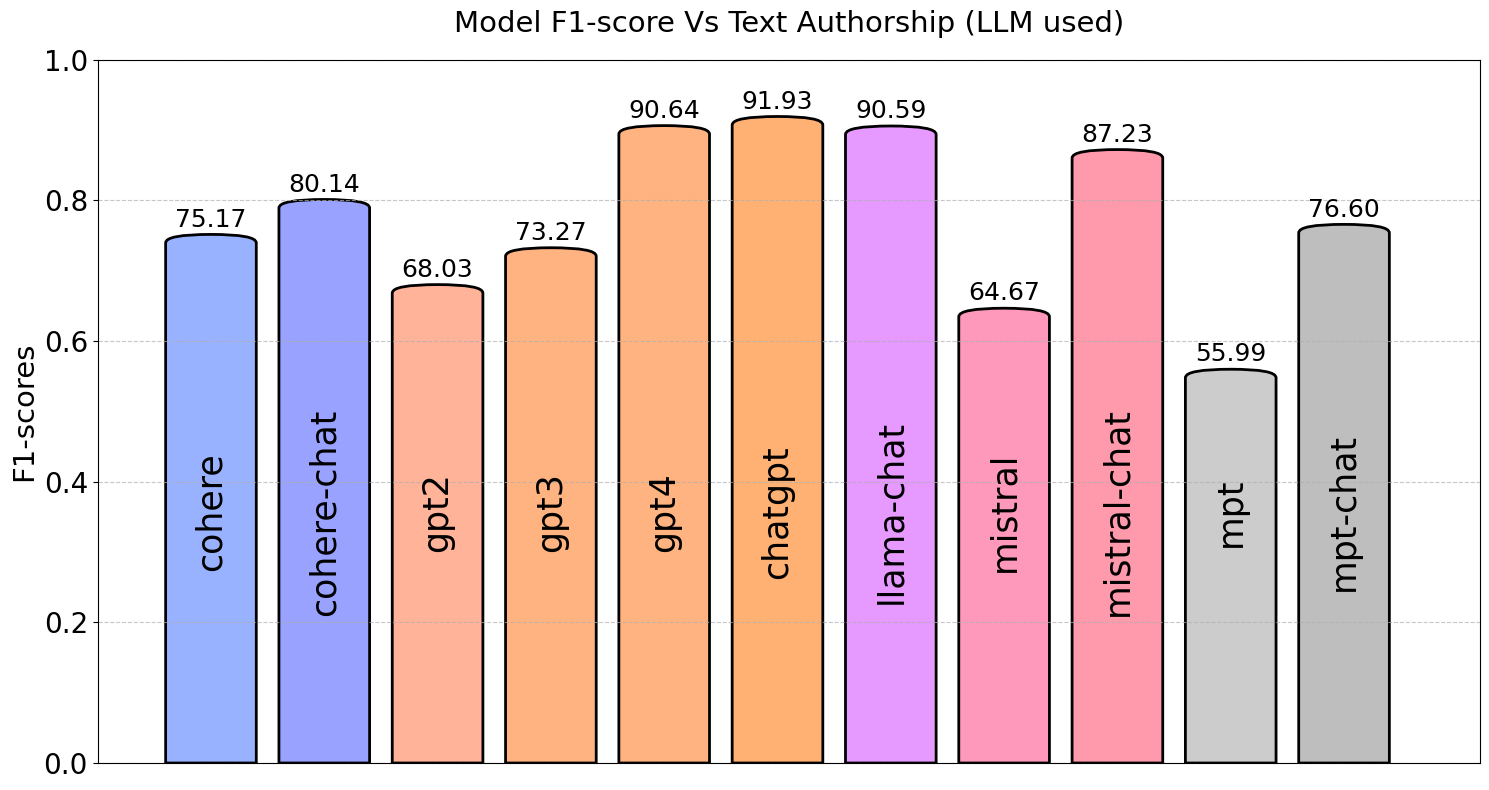}
    \caption{F1 scores VS the generator's texts : instruct/chat variants were harder to detect than the corresponding base models.}
    \label{figure:02}
\end{figure}

\begin{figure}[!h]
    \centering
    \includegraphics[width=1\linewidth]{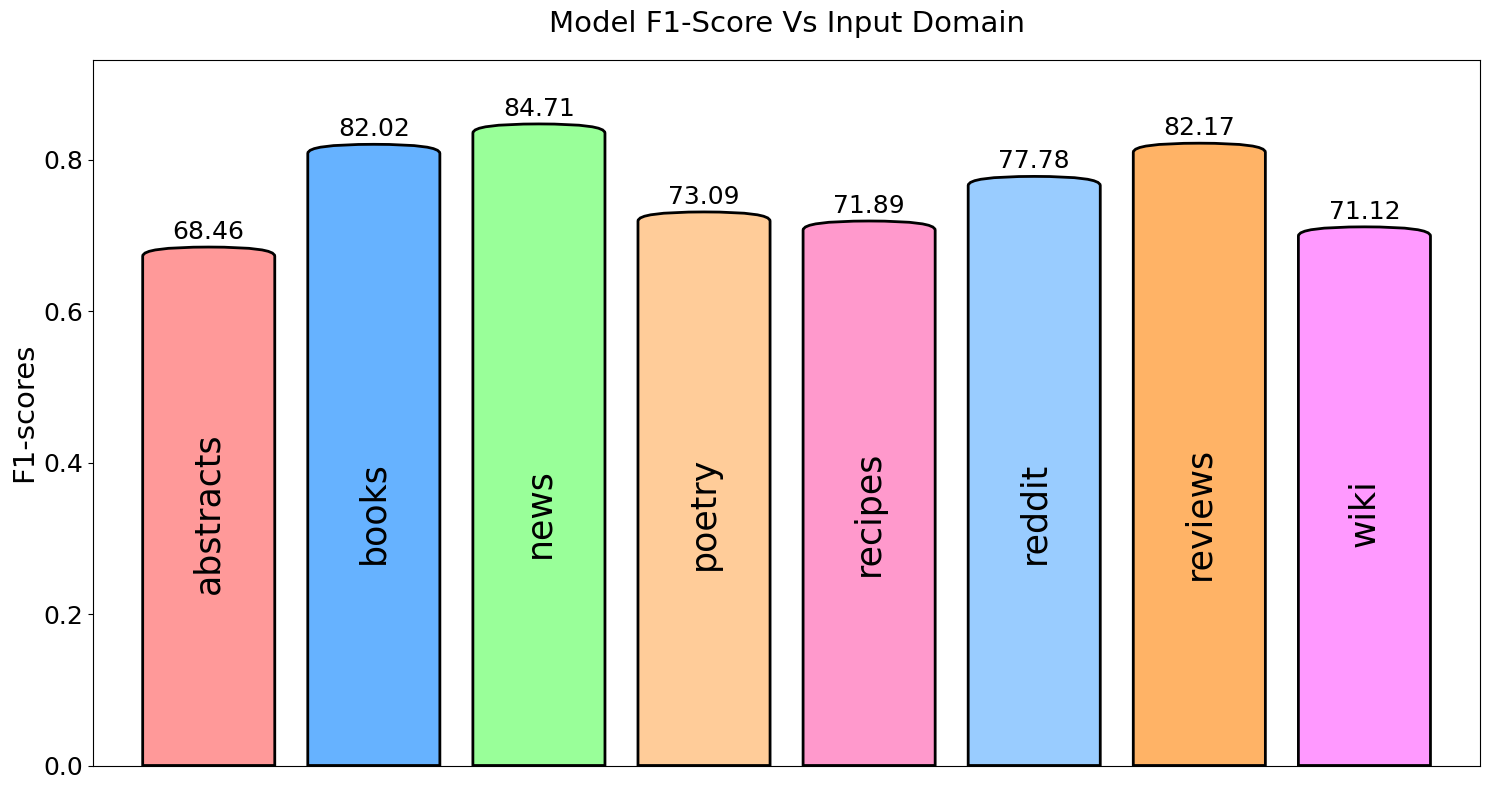}
    \caption{F1 scores VS each domain's texts : news i.e the only training data domain was easier to detect.}
    \label{figure:03}
\end{figure}

\begin{figure}[!h]
    \centering
    \includegraphics[width=1\linewidth]{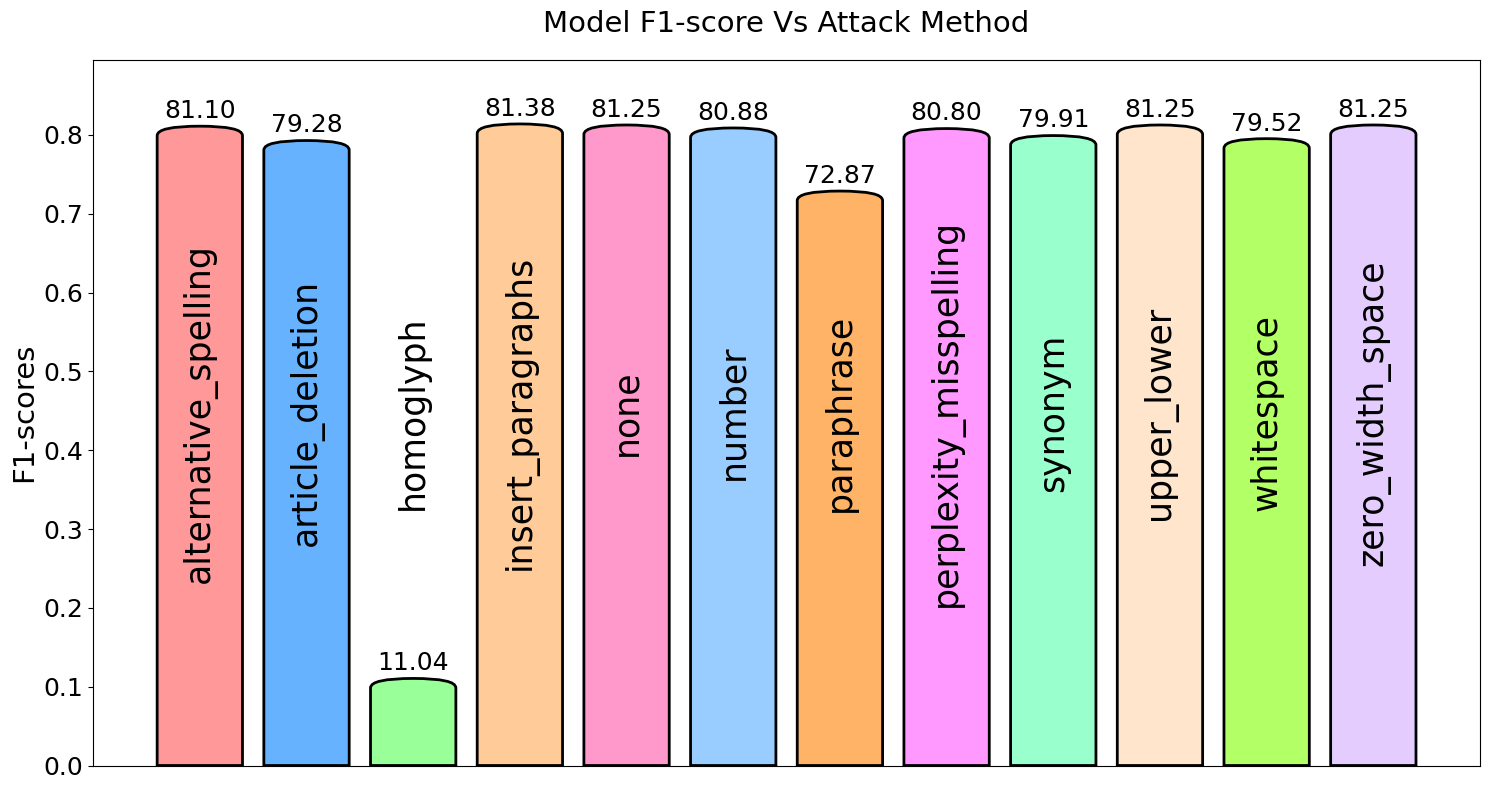}
    \caption{F1 scores VS adversarial method used in the input texts : homo-glyphs are the only real issue, and paraphrases to a small extent, while the rest can be handled through pre-processing.}
    \label{figure:04}
\end{figure}

We have also tested over raid-bench \citep{dugan-etal-2025-genai}, which consists of texts from 11 generators and 8 domains. Among them, roughly 10\% would be from a seen domain (news articles), while the rest are unseen by our models. The dataset's texts were also created using various sampling strategies (greedy, random, etc.). The texts were also modified to have adversarial methods, including homoglyphs, misspellings, alternative spellings, article deletion, etc. Among the 11 generators used, GPT-4 is one that is similar to the generator whose outputs our model has been trained on (GPT-4o). However, both of them have different linguistic and stylistic features, similar to how GPT-4 is different from GPT-3. We have tested our model's performance once again upon being trained on our own full training data. Additionally, we have also performed an error analysis to find out what domains, models, attack strategies, and decoding strategies affected the model's performance and to what extent. This can be seen in \autoref{figure:01}, \autoref{figure:02}, \autoref{figure:03}, and \autoref{figure:04}. The texts were classified as machine generated if at least one percent of the tokens within the model's context length were classified as machine generated. The F1 score obtained with the initial model trained on a single generator was 0.63, and the F1 score grew to 0.79 upon being trained on our full dataset. Evaluation was done directly without performing any preprocessing of the texts, and neither were our models trained on texts with any of the adversarial methods.
\subsection{Comparisons with other systems}
\label{sec:proprietary}
While many proprietary systems claim to have excellent results, they often struggle with unseen domains and generators. Further, many systems like ZeroGPT\footnote{\url{https://github.com/zerogpt-net/zerogpt-api}} and GPTZero\footnote{\url{https://gptzero.stoplight.io/docs/gptzero-api/}} do not provide fine-grained predictions through their API. This would require manually evaluating the results through the UI by counting correct and incorrectly classified word counts. Several users have tried this manually over a subset \citep{kadiyala-2024-rkadiyala} only to find a large gap in performance. However, for providing a comparison, we use other open-source systems for binary classification over the same benchmarks. 
\section{Other Observations}
\label{sec:findings}
The sentences inside which text authorship switches from human to LLM or vice versa were found to be relatively shorter than the original text portions that they replaced. LLMs may be likely to finish the current sentence earlier than usual to move on to the next sentence in text completion scenarios. The mean length of the original portion and the replaced portions of those sentences for each language and generator can be seen in \autoref{table:4} and \autoref{table:5}, respectively. This observation was consistent across all languages and generators with a 20-30\% reduction and a larger reduction in Hindi. For Chinese and Japanese too, we did observe a 20-30\% reduction in character count when comparing the original and replaced portions of the sentence after the text boundary. Although there is a good variation in this feature across languages, the mean and medians observed for each language were similar for all the LLMs. This is further elaborated in \autoref{sec:prepostboundary}. \\ 
\begin{table}[!h]
    \centering
    \begin{tabular}{ccc}
    \noalign{\hrule height 2pt}
    \rowcolor{cyan}\textbf{\small{Language}} & \textbf{\small{Length of}} & \textbf{\small{Length of}} \\
    \rowcolor{cyan}                  & \textbf{\small{Original part}} & \textbf{\small{generated part}} \\
    \noalign{\hrule height 2pt}     
    \small{Arabic}      & \small{17} & \small{13} \\
    \small{Czech}       & \small{11} &  \small{8} \\
    \small{Dutch}       & \small{12} & \small{10} \\
    \small{English}     & \small{15} & \small{11} \\
    \small{French}      & \small{14} & \small{11} \\
    \small{German}      & \small{12} &  \small{9} \\
    \small{Greek}       & \small{15} & \small{12} \\
    \small{Hebrew}      & \small{11} &  \small{9} \\
    \small{Hindi}       & \small{26} & \small{12} \\
    \small{Indonesian}  & \small{11} &  \small{8} \\
    \small{Italian}     & \small{15} & \small{14} \\
    \small{Korean}      & \small{9}  &  \small{7} \\
    \small{Persian}     & \small{17} & \small{15} \\
    \small{Polish}      & \small{10} &  \small{7} \\
    \small{Portuguese}  & \small{15} & \small{11} \\
    \small{Romanian}    & \small{14} & \small{11} \\
    \small{Russian}     & \small{11} &  \small{9} \\
    \small{Spanish}     & \small{15} & \small{12} \\
    \small{Turkish}     & \small{10} &  \small{8} \\
    \small{Ukrainian}   & \small{11} &  \small{8} \\
    \small{Vietnamese}  & \small{18} & \small{14} \\
    \noalign{\hrule height 2pt}  
    \rowcolor{grey} \textbf{\small{Average}}     & \textbf{\small{13.8}} & \textbf{\small{10.4}} \\ 
    \noalign{\hrule height 2pt}  
    \end{tabular}
    \caption{Median length (words) of original \& newly generated parts of the sentences - Language wise : Models tend to finish off current sentence after authorship switch quickly before continuing with the rest of the text.}
    \label{table:4}
\end{table}
\begin{table}[!h]
    \centering
    \begin{tabular}{ccc}
    \noalign{\hrule height 2pt}
    \rowcolor{cyan}\textbf{\small{Generator}} & \textbf{\small{Length of}}      & \textbf{\small{Length of}} \\
    \rowcolor{cyan}                   & \textbf{\small{original part}}  & \textbf{\small{generated part}} \\
    \noalign{\hrule height 2pt}     
    \small{Amazon-Nova-Pro}          & 14 & 10 \\
    \small{Amazon-Nova-Lite}         & 12 & 10 \\
    \small{Aya-23-35B}               & 11 & 10 \\
    \small{Claude-3.5-Haiku}         & 18 & 10 \\
    \small{Claude-3.5-Sonnet}        & 16 & 10 \\
    \small{Command-R-Plus}           & 16 & 10 \\
    \small{GPT-4o}                   & 12 & 10 \\
    \small{GPT-o1}                   & 11 &  9 \\
    \small{Gemini-1.5-Pro}           & 15 & 10 \\
    \small{Gemini-1.5-Flash}         &  9 & 10 \\
    \small{Mistral-Large-2411}       & 11 & 10 \\
    \small{Perplexity-Sonar-large}   & 15 & 11 \\
    \noalign{\hrule height 2pt} 
    \rowcolor{grey} \small{\textbf{Average}}      & 13.3 & 10 \\
    \noalign{\hrule height 2pt} 
    \end{tabular}
    \caption{Median length (words) of original \& newly generated parts of the sentences : Generator wise}
    \label{table:5}
\end{table}
\begin{figure*}[!ht]
    \centering
    \includegraphics[width=0.82\linewidth]{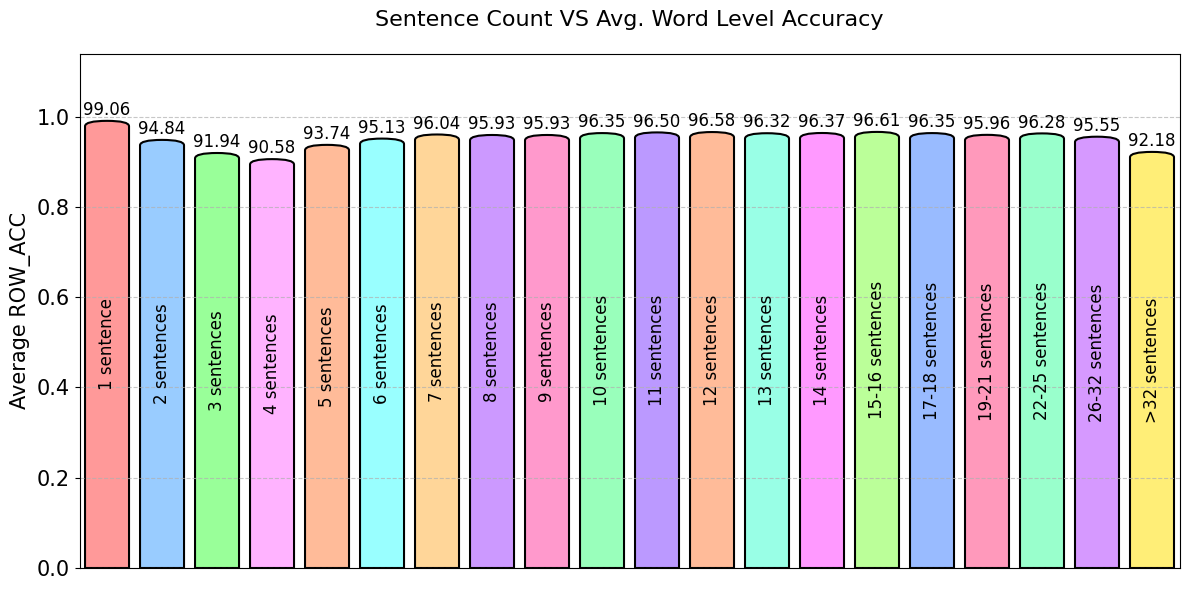}
    \caption{Accuracy VS length of input texts (sentence count)}
    \label{figure:05}
\end{figure*}
\begin{table}[!h]
    \centering
    \begin{tabular}{ccc}
    \noalign{\hrule height 2pt}
    \rowcolor{cyan}\textbf{\small{Generator}} & \textbf{\small{Accuracy}}\\
    \noalign{\hrule height 2pt}     
    \small{Amazon-Nova-Pro}          & \small{94.90} \\
    \small{Amazon-Nova-Lite}         & \small{95.26} \\
    \small{Aya-23-35B}               & \small{91.75} \\
    \small{Claude-3.5-Haiku}         & \small{96.07} \\
    \small{Claude-3.5-Sonnet}        & \small{95.97} \\
    \small{Command-R-Plus}           & \small{93.92} \\
    \small{GPT-4o}                   & \small{91.78} \\
    \small{GPT-o1}                   & \small{96.61} \\
    \small{Gemini-1.5-Flash}         & \small{92.34} \\
    \small{Gemini-1.5-Pro}           & \small{93.38} \\
    \small{Mistral-Large-2411}       & \small{93.47} \\
    \small{Perplexity-Sonar-large}   & \small{94.91} \\
    \noalign{\hrule height 2pt} 
    \rowcolor{grey} \textbf{\small{Average}}         & \small{94.31} \\
    \noalign{\hrule height 2pt} 
    \end{tabular}
    \caption{Word level accuracy (.2f) of our models over our dataset (English)}
    \label{table:6}
\end{table}
\section{Conclusion}
\label{sec:conclusion}
Despite not being trained on the domains or generators, the models built through our approach performed well over several benchmarks as seen in \autoref{subsec:etsessayssharedtask} and \autoref{subsec:raidbenchsharedtask} over inputs that were from non-native speakers and consist of adversarial methods. Further, one case where many proprietary systems struggle is when the inputs were too short, which our models were able to overcome as seen in \autoref{figure:05}, which demonstrates our models' accuracy over our test set compared to the input text's sentence count. \autoref{table:6} displays our model's performance over the English subset of our dataset for each generator. A similar trend from \autoref{subsec:raidbenchsharedtask} was observed with models that are likely less instruction-tuned or not instruction-tuned; they tend to produce texts that are harder to distinguish than their alternatives.
\subsection{Scalability and scope for extension}
\label{subsec:scalability}
The original dataset used to train our current models, as mentioned in \autoref{sec:dataset}, consists of samples in over 60 languages, which would cover 70\% of the world population's primary language, and all of the languages are supported by existing multilingual transformer models, making the process of scaling the work to more languages easier. Despite not being trained on the generators or domains' texts, our models were able to perform well on several benchmarks. They even reached an F1 score of 0.79 against adversarial inputs while they were neither trained over them nor pre-processed. Similarly, the creation and usage of such large datasets of other domains along with ours might result in robust and better models. We couldn't explore the relation between the instruction tuning sample size of LLMs and the detectability of their texts due to the proprietary nature of most of the generators we used, but a similar study using open-data models could uncover more insights. 
\subsection{Scope for Improvement}
\label{subsec:scopeforimprovement}
As seen in \autoref{figure:04}, almost none of the adversarial methods affected the models built through our approach other than paraphrasing and homoglyphs. However, homoglyphs can be pre-processed by mapping them to the actual character they were imitating in the text. This would require a large collection of homo-glyph-to-character mapping sets to use for preprocessing. Further, paraphrased samples of various numbers of iterations being included in the training dataset might lead to further improvements. It is also worth exploring how detectable texts are in cases where multiple generators contribute a portion each in a human-authored text. Other missing adversarial methods that are likely to be used in practical scenarios include the usage of proprietary systems that 'humanize' a given text in an attempt to evade detection.
\section*{Limitations}
\label{sec:limitations}
Just like any other detector or classifier, no detector can guarantee a 100\% accuracy and hence the models are not meant to be used directly for decision making but are meant to be used in a human-in-the-loop scenarios. Furthermore, the experiments carried out did not include cases of multiple LLMs co-authoring a portion each of the same text. The models were built primarily for a human-in-the-loop use cases where the model would try to flag most of the likely machine-generated portions while the flagged content can be validated either through an ensemble of models or a human and hence a tilt towards higher recall can be observed in the metrics
\bibliography{anthology,custom}
\bibliographystyle{acl_natbib}
\appendix
\newpage
\section{Pre- and Post- Boundary Comparisons}
\label{sec:prepostboundary}
The mean and median word counts of the text portions in a sentence after the text authorship shifts from human to LLM can be seen in \autoref{table:101} and \autoref{table:102} in comparison to the texts they replace.
\section{Dataset Creation}
\label{suc:datasetreproducibility}
The max\_new\_tokens value specified to the generator during creation of partial cases was randomized between 80\% to 200\% of the length of the portion that is being replaced. The prompts used for creation of the partial samples and rewritten samples can be seen in \autoref{table:401} and \autoref{table:402} respectively.
\begin{table}[!h]
    \centering
    \begin{tabular}{|c|}
    \hline
    \small{continue this text in \underline{Language} directly : }          \\
    \small{complete this text in \underline{Language}, respond directly : } \\
    \hline
    \end{tabular}
    \caption{Prompts used in dataset creation : Partial cases}
    \label{table:401}
\end{table}
\begin{table}[!h]
    \centering
    \begin{tabular}{|c|}
    \hline
    \small{Rewrite this in \underline{Language} a different way : }            \\
    \small{Generate an alternative version of this in \underline{Language} : } \\
    \small{Generate a later update to this in \underline{Language} : }         \\
    \small{Generate a previous version of this in \underline{Language} ; }     \\
    \hline
    \end{tabular}
    \caption{Prompts used in dataset creation : Rewritten cases}
    \label{table:402}
\end{table}
\begin{table}[!h]
    \centering
    \begin{tabular}{ccc}
    \noalign{\hrule height 2pt}
    \rowcolor{cyan} \textbf{Hyperparameter}  & \textbf{Value} \\
    \noalign{\hrule height 2pt}     
        Seed (Training)                 & 1024 \\
        Seed (Shuffling)                & 1024 \\
        Number of Epochs                & 5  \\
        Per Device Batch Size (Train)   & 12 \\
        Per Device Batch Size (Eval)    & 30 \\
        Context Length                  & 16384 \\
        Learning Rate                   & 5e-5 \\
        Weight Decay                    & 0 \\
        Dropout (CRF Layer)             & 0.075 \\
    \noalign{\hrule height 2pt}     
    \end{tabular}
    \caption{Training Hyper-parameters used}
    \label{table:301}
\end{table}
\begin{table*}[!h]
    \centering
    \begin{tabular}{ccccc}
    \noalign{\hrule height 2pt}
    \rowcolor{cyan}\textbf{Language }$\downarrow$ & \textbf{Mean length of}   & \textbf{Mean length of}     & \textbf{Median length of} & \textbf{Median length of} \\
    \rowcolor{cyan}                               & \textbf{old text portion} & \textbf{new text portion}   & \textbf{Old text portion} & \textbf{New text portion} \\
    \noalign{\hrule height 2pt}
    Arabic            & 18.73 & 16.25 & 17 & 13 \\
    Czech             & 12.02 &  9.52 & 11 & 8  \\
    Dutch             & 13.73 & 13.39 & 12 & 10 \\
    English           & 16.04 & 14.59 & 15 & 11 \\
    French            & 15.50 & 13.16 & 14 & 11 \\
    German            & 13.03 & 10.89 & 12 & 9  \\
    Greek             & 16.74 & 14.87 & 15 & 12 \\
    Hebrew            & 12.64 & 10.65 & 11 & 9  \\
    Hindi             & 40.56 & 15.42 & 26 & 12 \\
    Indonesian        & 12.44 &  9.56 & 11 & 8  \\
    Italian           & 17.54 & 16.39 & 15 & 14 \\
    Korean            &  9.85 &  8.08 &  9 & 7  \\
    Persian           & 18.83 & 19.88 & 17 & 15 \\
    Polish            & 11.42 &  8.84 & 10 & 7  \\
    Portuguese        & 16.52 & 13.29 & 15 & 11 \\
    Romanian          & 16.30 & 13.50 & 14 & 11 \\
    Russian           & 12.27 & 10.63 & 11 & 9  \\
    Spanish           & 17.18 & 14.81 & 15 & 12 \\
    Turkish           & 11.81 &  9.74 & 10 & 8  \\
    Ukrainian         & 12.04 & 10.39 & 11 & 8  \\
    Vietnamese        & 20.06 & 18.01 & 18 & 14 \\
    \noalign{\hrule height 2pt}
    \rowcolor{grey} \textbf{Average}  & \textbf{16.19} & \textbf{12.95} & \textbf{13.76} & \textbf{10.43} \\
    \noalign{\hrule height 2pt}     
    \end{tabular}
    \caption{Comparison of replaced and generated text portion lengths (word count) : Language wise}
    \label{table:101}
\end{table*}
\begin{table*}[!h]
    \centering
    \begin{tabular}{ccccc}
    \noalign{\hrule height 2pt}
    \rowcolor{cyan}\textbf{Generator}$\downarrow$ & \textbf{Mean length of}   & \textbf{Mean length of}     & \textbf{Median length of} & \textbf{Median length of} \\
    \rowcolor{cyan}                               & \textbf{old text portion} & \textbf{new text portion}   & \textbf{Old text portion} & \textbf{New text portion} \\
    \noalign{\hrule height 2pt}     
    \small{Amazon-Nova-Pro}          & 16.02 & 13.27 & 12 & 10 \\
    \small{Amazon-Nova-Lite}         & 18.12 & 12.87 & 14 & 10 \\
    \small{Aya-23-35B}               & 13.70 & 12.87 & 11 & 10 \\
    \small{Claude-3.5-Haiku}         & 20.19 & 13.13 & 18 & 10 \\
    \small{Claude-3.5-Sonnet}        & 17.32 & 12.98 & 16 & 10 \\
    \small{Command-R-Plus}           & 16.92 & 13.28 & 16 & 10 \\
    \small{GPT-4o}                   & 13.49 & 12.84 & 12 & 10 \\
    \small{GPT-o1}                   & 14.83 & 12.36 & 11 & 9 \\
    \small{Gemini-1.5-Flash}         & 19.68 & 13.44 & 15 & 10 \\
    \small{Gemini-1.5-pro}           & 12.50 & 13.48 & 9 & 10 \\
    \small{Mistral-Large-2411}       & 12.85 & 12.85 & 11 & 10 \\
    \small{Perplexity-Sonar-large}   & 17.13 & 13.45 & 15 & 11 \\
    \noalign{\hrule height 2pt} 
    \rowcolor{grey} \textbf{Average}                 & \textbf{16.06} & \textbf{13.07} & \textbf{13.33} & \textbf{10} \\
    \noalign{\hrule height 2pt} 
    \end{tabular}
    \caption{Comparison of replaced and generated text portion lengths (word count) : Generator wise}
    \label{table:102}
\end{table*}
\section{Reproducibility}
\label{sec:reproducibility}
We used multilingual longformer \footnote{\url{https://huggingface.co/hyperonym/xlm-roberta-longformer-base-16384}} with an additional CRF layer. The hyper-parameters used for training the models can be seen in \autoref{table:301}. We built a separate model for each language, the training was done over H100 SXM over 237 hours. 
\section{Other Metrics}
\label{sec:otherresults}
The metrics over each type of text for each language and LLM separately can be seen in \autoref{table:103} to \autoref{table:114} respectively..
\begin{table*}[!h]
    \centering
    \begin{tabular}{ccccc}
    \noalign{\hrule height 2pt}
    \rowcolor{cyan}\textbf{Language }$\downarrow$ & \textbf{Accuracy} & \textbf{Precision} & \textbf{Recall} & \textbf{F1-score} \\
    \noalign{\hrule height 2pt}
    Arabic            & 96.44 & 92.50 & 97.17 & 94.78 \\
    Chinese*          & 86.58 & 87.03 & 86.46 & 86.75 \\
    Czech             & 94.98 & 94.57 & 97.96 & 96.23 \\
    Dutch             & 95.31 & 93.34 & 97.97 & 95.60 \\
    English           & 96.02 & 92.34 & 98.44 & 95.29 \\
    French            & 94.91 & 93.64 & 98.42 & 95.97 \\
    German            & 95.28 & 94.87 & 98.38 & 96.59 \\
    Greek             & 94.37 & 93.69 & 96.51 & 95.08 \\
    Hebrew            & 95.70 & 95.32 & 97.94 & 96.61 \\
    Hindi             & 96.34 & 89.72 & 96.66 & 93.06 \\
    Indonesian        & 96.64 & 95.61 & 98.29 & 96.93 \\
    Italian           & 95.38 & 95.04 & 97.58 & 96.29 \\
    Japanese*         & 86.13 & 85.64 & 94.17 & 89.70 \\
    Korean            & 95.77 & 95.29 & 97.69 & 96.48 \\
    Persian           & 94.36 & 84.45 & 96.88 & 90.24 \\
    Polish            & 95.94 & 96.76 & 97.19 & 96.97 \\
    Portuguese        & 94.89 & 91.92 & 96.07 & 93.95 \\
    Romanian          & 96.10 & 95.81 & 98.53 & 97.15 \\
    Russian           & 94.02 & 86.67 & 97.29 & 91.67 \\
    Spanish           & 94.47 & 90.02 & 98.14 & 93.90 \\
    Turkish           & 93.62 & 88.56 & 97.17 & 92.66 \\
    Ukrainian         & 93.53 & 86.58 & 97.93 & 91.90 \\
    Vietnamese        & 89.67 & 77.23 & 97.44 & 86.17 \\
    \noalign{\hrule height 2pt}
    \rowcolor{grey} \textbf{Average}  & \textbf{94.19} & \textbf{91.16} & \textbf{96.97} & \textbf{93.91} \\
    \noalign{\hrule height 2pt}     
    \end{tabular}
    \caption{Word-level Metrics of our models over each language : our test set}
    \caption*{* Character level evaluations were done instead for Japanese and Chinese}
    \label{table:115}
\end{table*}
\begin{table*}[!h]
    \centering
    \small{
    \begin{tabular}{ccccc}
    \noalign{\hrule height 2pt}
    \rowcolor{cyan}\textbf{Language }$\downarrow$ & \textbf{Partial cases} & \textbf{Unchanged cases} & \textbf{Rewritten cases} & \textbf{Overall} \\
    \noalign{\hrule height 2pt}
    Arabic            & 97.56 & 88.10 & 98.09 & 97.17 \\
    Chinese*          & 93.70 & 75.35 & 91.60 & 87.00 \\
    Czech             & 96.63 & 80.10 & 94.36 & 95.20 \\
    Dutch             & 95.21 & 78.00 & 92.10 & 95.23 \\
    English           & 97.77 & 89.61 & 98.87 & 96.60 \\
    French            & 97.34 & 72.85 & 97.14 & 95.28 \\
    German            & 96.92 & 75.73 & 95.29 & 95.58 \\
    Greek             & 95.65 & 81.80 & 82.85 & 92.96 \\
    Hebrew            & 97.35 & 70.89 & 96.26 & 95.73 \\
    Hindi             & 96.65 & 92.82 & 96.67 & 96.59 \\
    Indonesian        & 97.27 & 85.73 & 95.76 & 96.60 \\
    Italian           & 96.88 & 80.88 & 94.70 & 95.65 \\
    Japanese*         & 97.48 & 88.57 & 93.94 & 96.85 \\
    Korean            & 97.68 & 84.15 & 93.73 & 95.39 \\
    Persian           & 96.91 & 89.45 & 93.22 & 94.05 \\
    Polish            & 96.96 & 87.52 & 92.32 & 95.98 \\
    Portuguese        & 95.28 & 94.15 & 96.32 & 95.28 \\
    Romanian          & 96.53 & 76.55 & 96.64 & 96.23 \\
    Russian           & 96.46 & 79.10 & 94.45 & 94.03 \\
    Spanish           & 96.92 & 71.75 & 96.97 & 94.97 \\
    Turkish           & 95.55 & 82.68 & 98.17 & 92.65 \\
    Ukrainian         & 95.39 & 73.81 & 95.48 & 93.90 \\
    Vietnamese        & 94.46 & 76.17 & 97.14 & 88.74 \\
    \noalign{\hrule height 2pt}
    \end{tabular}
    }
    \caption{Case wise accuracies over all languages for each generator : amazon-nova-pro}
    \label{table:103}
\end{table*}
\begin{table*}[!h]
    \centering
    \small{
    \begin{tabular}{ccccc}
    \noalign{\hrule height 2pt}
    \rowcolor{cyan}\textbf{Language }$\downarrow$ & \textbf{Partial cases} & \textbf{Unchanged cases} & \textbf{Rewritten cases} & \textbf{Overall} \\
    \noalign{\hrule height 2pt}
    Arabic            & 96.56 & 90.93 & 95.62 & 95.88 \\
    Chinese*          & 93.20 & 76.99 & 93.57 & 87.77 \\
    Czech             & 97.81 & 79.40 & 94.58 & 96.43 \\
    Dutch             & 97.07 & 78.10 & 92.85 & 95.19 \\
    English           & 98.11 & 89.25 & 98.73 & 96.80 \\
    French            & 97.59 & 76.28 & 97.79 & 96.07 \\
    German            & 98.01 & 76.34 & 95.52 & 96.76 \\
    Greek             & 96.00 & 79.78 & 88.01 & 93.66 \\
    Hebrew            & 98.05 & 83.84 & 94.35 & 96.78 \\
    Hindi             & 96.49 & 91.59 & 95.30 & 95.38 \\
    Indonesian        & 97.99 & 85.03 & 97.18 & 97.06 \\
    Italian           & 96.95 & 80.81 & 95.45 & 95.54 \\
    Japanese*         & 98.07 & 76.50 & 93.02 & 92.78 \\
    Korean            & 98.20 & 82.49 & 95.42 & 95.68 \\
    Persian           & 97.40 & 88.48 & 94.92 & 95.31 \\
    Polish            & 97.55 & 89.32 & 93.83 & 96.63 \\
    Portuguese        & 92.67 & 87.92 & 95.09 & 94.35 \\
    Romanian          & 97.97 & 76.44 & 93.76 & 96.20 \\
    Russian           & 97.22 & 81.47 & 96.30 & 95.10 \\
    Spanish           & 97.49 & 71.55 & 97.15 & 94.98 \\
    Turkish           & 96.63 & 83.99 & 90.84 & 93.87 \\
    Ukrainian         & 96.74 & 74.80 & 99.91 & 94.24 \\
    Vietnamese        & 95.28 & 78.87 & 97.03 & 89.76 \\
    \noalign{\hrule height 2pt}
    \end{tabular}
    }
    \caption{Case wise accuracies over all languages for each generator : amazon-nova-lite}
    \label{table:104}
\end{table*}
\begin{table*}[!h]
    \centering
    \small{
    \begin{tabular}{ccccc}
    \noalign{\hrule height 2pt}
    \rowcolor{cyan}\textbf{Language }$\downarrow$ & \textbf{Partial cases} & \textbf{Unchanged cases} & \textbf{Rewritten cases} & \textbf{Overall} \\
    \noalign{\hrule height 2pt}
    Arabic            & 95.14 & 92.22 & 97.76 & 96.05 \\
    Chinese*          & 85.65 & 75.66 & 89.10 & 82.40 \\
    Czech             & 89.75 & 79.02 & 87.94 & 89.24 \\
    Dutch             & 92.45 & 79.97 & 92.99 & 93.12 \\
    English           & 93.01 & 90.49 & 96.96 & 93.52 \\
    French            & 93.28 & 73.12 & 95.14 & 92.72 \\
    German            & 89.89 & 77.28 & 92.53 & 90.06 \\
    Greek             & 92.04 & 80.69 & 91.83 & 91.75 \\
    Hebrew            & 96.71 & 82.75 & 91.54 & 95.32 \\
    Hindi             & 94.18 & 93.62 & 92.96 & 94.88 \\
    Indonesian        & 90.91 & 83.55 & 95.28 & 92.85 \\
    Italian           & 89.21 & 75.43 & 88.87 & 88.87 \\
    Japanese*         & 75.56 & 78.10 & 91.21 & 75.64 \\
    Korean            & 95.04 & 85.46 & 92.92 & 94.14 \\
    Persian           & 93.81 & 87.28 & 95.29 & 92.98 \\
    Polish            & 90.40 & 86.41 & 89.15 & 90.81 \\
    Portuguese        & 92.69 & 91.17 & 90.96 & 92.69 \\
    Romanian          & 93.65 & 78.15 & 95.16 & 93.17 \\
    Russian           & 93.00 & 79.77 & 92.12 & 92.20 \\
    Spanish           & 91.30 & 72.87 & 93.17 & 91.88 \\
    Turkish           & 90.19 & 82.59 & 98.19 & 90.77 \\
    Ukrainian         & 87.77 & 73.69 & 97.57 & 90.69 \\
    Vietnamese        & 87.70 & 76.08 & 96.83 & 88.54 \\
    \noalign{\hrule height 2pt}
    \end{tabular}
     }
   \caption{Case wise accuracies over all languages for each generator : Aya-23}
    \label{table:105}
\end{table*}
\begin{table*}[!h]
    \centering
    \small{
    \begin{tabular}{ccccc}
    \noalign{\hrule height 2pt}
    \rowcolor{cyan}\textbf{Language }$\downarrow$ & \textbf{Partial cases} & \textbf{Unchanged cases} & \textbf{Rewritten cases} & \textbf{Overall} \\
    \noalign{\hrule height 2pt}
    Arabic            & 98.82 & 91.63 & 95.75 & 97.18 \\
    Chinese*          & 86.51 & 75.93 & 86.18 & 87.75 \\
    Czech             & 99.80 & 80.78 & 91.56 & 97.55 \\
    Dutch             & 99.49 & 77.57 & 86.16 & 96.41 \\
    English           & 99.37 & 90.98 & 98.32 & 97.76 \\
    French            & 99.63 & 74.86 & 90.15 & 96.30 \\
    German            & 99.79 & 77.23 & 94.62 & 97.37 \\
    Greek             & 99.90 & 87.33 & 82.84 & 97.46 \\
    Hebrew            & 98.94 & 83.48 & 82.61 & 96.27 \\
    Hindi             & 98.72 & 92.35 & 95.48 & 97.23 \\
    Indonesian        & 99.55 & 88.19 & 94.11 & 98.05 \\
    Italian           & 99.97 & 81.43 & 93.49 & 97.48 \\
    Japanese*         & 98.33 & 87.97 & 91.08 & 97.02 \\
    Korean            & 99.40 & 84.22 & 94.45 & 96.93 \\
    Persian           & 97.99 & 89.54 & 90.00 & 94.54 \\
    Polish            & 99.60 & 88.75 & 85.69 & 97.62 \\
    Portuguese        & 99.17 & 90.82 & 82.19 & 96.52 \\
    Romanian          & 99.93 & 78.70 & 92.11 & 97.11 \\
    Russian           & 99.26 & 80.44 & 92.17 & 95.36 \\
    Spanish           & 99.31 & 71.65 & 93.24 & 95.91 \\
    Turkish           & 98.60 & 81.65 & 92.86 & 94.62 \\
    Ukrainian         & 99.27 & 73.52 & 91.46 & 93.99 \\
    Vietnamese        & 98.42 & 77.05 & 92.41 & 91.85 \\
    \noalign{\hrule height 2pt}
    \end{tabular}
    }
    \caption{Case wise accuracies over all languages for each generator : Claude-3.5-Haiku}
    \label{table:106}
\end{table*}
\begin{table*}[!h]
    \centering
    \small{
    \begin{tabular}{ccccc}
    \noalign{\hrule height 2pt}
    \rowcolor{cyan}\textbf{Language }$\downarrow$ & \textbf{Partial cases} & \textbf{Unchanged cases} & \textbf{Rewritten cases} & \textbf{Overall} \\
    \noalign{\hrule height 2pt}
    Arabic            & 98.63 & 91.82 & 100.00 & 96.58 \\
    Chinese*          & 92.64 & 78.36 & 95.14 & 88.43 \\
    Czech             & 99.30 & 77.22 & 99.88 & 97.62 \\
    Dutch             & 99.30 & 77.53 & 99.52 & 97.30 \\
    English           & 99.35 & 90.02 & 99.76 & 98.03 \\
    French            & 99.53 & 73.66 & 99.97 & 97.06 \\
    German            & 99.52 & 76.06 & 99.69 & 97.33 \\
    Greek             & 99.00 & 80.60 & 99.62 & 95.83 \\
    Hebrew            & 97.68 & 82.69 & 99.88 & 96.46 \\
    Hindi             & 99.12 & 92.51 & 99.88 & 97.63 \\
    Indonesian        & 99.66 & 84.55 & 100.00 & 98.43 \\
    Italian           & 99.69 & 81.13 & 99.99 & 98.13 \\
    Japanese*         & 98.59 & 87.36 & 99.64 & 98.04 \\
    Korean            & 98.77 & 83.49 & 99.87 & 97.15 \\
    Persian           & 98.35 & 87.92 & 99.97 & 96.01 \\
    Polish            & 99.00 & 90.30 & 99.26 & 98.20 \\
    Portuguese        & 98.74 & 89.65 & 83.74 & 96.38 \\
    Romanian          & 99.18 & 80.36 & 99.71 & 97.46 \\
    Russian           & 99.33 & 80.55 & 99.93 & 94.55 \\
    Spanish           & 99.06 & 71.68 & 99.92 & 96.65 \\
    Turkish           & 98.45 & 83.13 & 99.96 & 95.32 \\
    Ukrainian         & 99.06 & 74.14 & 99.87 & 95.62 \\
    Vietnamese        & 98.10 & 77.40 & 99.92 & 88.87 \\
    \noalign{\hrule height 2pt}
    \end{tabular}
    }
    \caption{Case wise accuracies over all languages for each generator : Claude-3.5-Sonnet}
    \label{table:107}
\end{table*}
\begin{table*}[!h]
    \centering
    \small{
    \begin{tabular}{ccccc}
    \noalign{\hrule height 2pt}
    \rowcolor{cyan}\textbf{Language }$\downarrow$ & \textbf{Partial cases} & \textbf{Unchanged cases} & \textbf{Rewritten cases} & \textbf{Overall} \\
    \noalign{\hrule height 2pt}
    Arabic            & 87.12 & 85.34 & 86 & 82 \\
    Chinese*          & 88.90 & 86.45 & 89 & 84 \\
    English           & 92.45 & 90.12 & 91 & 88 \\
    French            & 89.78 & 87.21 & 90 & 85 \\
    German            & 90.23 & 88.05 & 89 & 86 \\
    Italian           & 89.12 & 87.00 & 89 & 86 \\
    Japanese*         & 87.77 & 85.88 & 88 & 83 \\
    Korean            & 88.56 & 86.34 & 87 & 85 \\
    Portuguese        & 90.12 & 88.34 & 89 & 87 \\
    Spanish           & 90.45 & 88.12 & 89 & 87 \\
    \noalign{\hrule height 2pt}
    \end{tabular}
    }
    \caption{Case wise accuracies over all languages for each generator : Command-R-Plus}
    \label{table:108}
\end{table*}
\begin{table*}[!h]
    \centering
    \small{
    \begin{tabular}{ccccc}
    \noalign{\hrule height 2pt}
    \rowcolor{cyan}\textbf{Language }$\downarrow$ & \textbf{Partial cases} & \textbf{Unchanged cases} & \textbf{Rewritten cases} & \textbf{Overall} \\
    \noalign{\hrule height 2pt}
    Arabic            & 95.74 & 91.26 & 96.31 & 95.01 \\
    Chinese*          & 92.63 & 77.87 & 92.51 & 86.36 \\
    Czech             & 93.30 & 80.96 & 91.67 & 91.87 \\
    Dutch             & 94.14 & 74.85 & 90.84 & 92.58 \\
    English           & 94.94 & 90.02 & 92.84 & 94.01 \\
    French            & 92.87 & 75.18 & 93.90 & 90.89 \\
    German            & 93.67 & 75.82 & 93.99 & 91.97 \\
    Greek             & 94.22 & 81.18 & 95.67 & 92.11 \\
    Hebrew            & 92.60 & 82.51 & 95.10 & 91.85 \\
    Hindi             & 96.56 & 92.44 & 96.95 & 96.16 \\
    Indonesian        & 95.08 & 84.88 & 95.88 & 94.70 \\
    Italian           & 93.35 & 79.95 & 92.72 & 92.72 \\
    Japanese*         & 93.98 & 88.44 & 94.19 & 93.84 \\
    Korean            & 94.44 & 84.84 & 93.09 & 93.61 \\
    Persian           & 94.83 & 88.32 & 94.68 & 93.34 \\
    Polish            & 94.53 & 89.51 & 89.36 & 93.73 \\
    Portuguese        & 95.50 & 88.58 & 85.07 & 93.93 \\
    Romanian          & 94.59 & 77.44 & 92.73 & 93.26 \\
    Russian           & 92.90 & 80.17 & 97.34 & 92.61 \\
    Spanish           & 93.54 & 69.64 & 91.87 & 92.13 \\
    Turkish           & 92.83 & 83.80 & 88.09 & 91.37 \\
    Ukrainian         & 91.69 & 74.93 & 96.81 & 90.33 \\
    Vietnamese        & 92.10 & 77.39 & 93.32 & 88.25 \\
    \noalign{\hrule height 2pt}
    \end{tabular}
     }
   \caption{Case wise accuracies over all languages for each generator : GPT-4o}
    \label{table:109}
\end{table*}
\begin{table*}[!h]
    \centering
    \small{
    \begin{tabular}{ccccc}
    \noalign{\hrule height 2pt}
    \rowcolor{cyan}\textbf{Language }$\downarrow$ & \textbf{Partial cases} & \textbf{Unchanged cases} & \textbf{Rewritten cases} & \textbf{Overall} \\
    \noalign{\hrule height 2pt}
    Arabic            & 99.10 & 90.46 & 97.08 & 97.92 \\
    Chinese*          & 95.08 & 76.01 & 86.18 & 87.30 \\
    Czech             & 98.84 & 80.76 & 86.30 & 97.05 \\
    Dutch             & 98.80 & 77.47 & 89.03 & 97.12 \\
    English           & 99.07 & 88.92 & 94.25 & 96.91 \\
    French            & 98.77 & 76.17 & 91.74 & 96.53 \\
    German            & 98.92 & 76.66 & 89.69 & 97.12 \\
    Greek             & 98.87 & 81.60 & 85.68 & 97.05 \\
    Hebrew            & 98.97 & 83.94 & 97.33 & 98.10 \\
    Hindi             & 99.10 & 92.12 & 97.42 & 97.33 \\
    Indonesian        & 98.96 & 84.98 & 99.00 & 98.45 \\
    Italian           & 97.04 & 80.93 & 99.10 & 96.16 \\
    Japanese*         & 90.78 & 73.63 & 85.24 & 78.08 \\
    Korean            & 99.16 & 83.18 & 83.13 & 97.09 \\
    Persian           & 98.72 & 87.01 & 94.43 & 95.50 \\
    Polish            & 99.04 & 90.29 & 86.92 & 97.86 \\
    Portuguese        & 98.65 & 88.47 & 83.50 & 95.18 \\
    Romanian          & 98.77 & 76.50 & 98.37 & 97.57 \\
    Russian           & 98.98 & 78.13 & 87.02 & 95.23 \\
    Spanish           & 98.80 & 71.70 & 92.79 & 96.12 \\
    Turkish           & 98.94 & 82.37 & 87.33 & 96.55 \\
    Ukrainian         & 99.05 & 75.07 & 93.34 & 96.45 \\
    Vietnamese        & 98.29 & 78.87 & 91.75 & 92.24 \\
    \noalign{\hrule height 2pt}
    \end{tabular}
    }
    \caption{Case wise accuracies over all languages for each generator : GPT-o1}
    \label{table:110}
\end{table*}
\begin{table*}[!h]
    \centering
    \small{
    \begin{tabular}{ccccc}
    \noalign{\hrule height 2pt}
    \rowcolor{cyan}\textbf{Language }$\downarrow$ & \textbf{Partial cases} & \textbf{Unchanged cases} & \textbf{Rewritten cases} & \textbf{Overall} \\
    \noalign{\hrule height 2pt}
    Arabic            & 93.90 & 88.40 & 98.59 & 94.86 \\
    Chinese*          & 89.29 & 75.98 & 94.92 & 85.09 \\
    Czech             & 91.43 & 78.82 & 97.15 & 92.04 \\
    Dutch             & 95.08 & 78.10 & 91.91 & 94.86 \\
    English           & 96.57 & 91.03 & 97.34 & 94.64 \\
    French            & 95.92 & 76.69 & 98.72 & 94.04 \\
    German            & 96.22 & 78.67 & 98.80 & 95.15 \\
    Greek             & 92.04 & 84.32 & 98.56 & 93.52 \\
    Hebrew            & 91.90 & 69.05 & 98.50 & 92.51 \\
    Hindi             & 95.55 & 93.52 & 98.72 & 95.66 \\
    Indonesian        & 96.84 & 83.45 & 98.55 & 95.92 \\
    Italian           & 94.70 & 76.47 & 97.03 & 94.89 \\
    Japanese*         & 84.59 & 87.53 & 96.26 & 87.86 \\
    Korean            & 94.54 & 85.07 & 99.41 & 94.66 \\
    Persian           & 95.68 & 89.47 & 96.54 & 94.66 \\
    Polish            & 93.51 & 88.07 & 97.34 & 94.60 \\
    Portuguese        & 95.22 & 88.75 & 94.90 & 94.28 \\
    Romanian          & 97.31 & 75.73 & 97.53 & 96.30 \\
    Russian           & 94.96 & 78.53 & 99.40 & 92.82 \\
    Spanish           & 95.30 & 74.56 & 99.47 & 94.09 \\
    Turkish           & 94.82 & 82.44 & 96.67 & 92.51 \\
    Ukrainian         & 89.37 & 73.96 & 98.39 & 91.68 \\
    Vietnamese        & 91.06 & 80.25 & 99.45 & 87.71 \\
    \noalign{\hrule height 2pt}
    \end{tabular}
    }
    \caption{Case wise accuracies over all languages for each generator : Gemini-1.5-Pro}
    \label{table:111}
\end{table*}
\begin{table*}[!h]
    \centering
    \small{
    \begin{tabular}{ccccc}
    \noalign{\hrule height 2pt}
    \rowcolor{cyan}\textbf{Language }$\downarrow$ & \textbf{Partial cases} & \textbf{Unchanged cases} & \textbf{Rewritten cases} & \textbf{Overall} \\
    \noalign{\hrule height 2pt}
    Arabic            & 98.55 & 89.55 & 98.00 & 95.88 \\
    Chinese*          & 89.94 & 75.81 & 94.03 & 87.75 \\
    Czech             & 98.45 & 80.42 & 99.78 & 96.67 \\
    Dutch             & 98.22 & 78.00 & 99.39 & 95.86 \\
    English           & 97.92 & 90.02 & 99.39 & 95.26 \\
    French            & 98.10 & 73.66 & 99.94 & 95.59 \\
    German            & 98.71 & 74.75 & 99.58 & 96.74 \\
    Greek             & 98.96 & 85.16 & 98.90 & 98.09 \\
    Hebrew            & 97.64 & 82.90 & 99.71 & 96.82 \\
    Hindi             & 98.17 & 93.06 & 99.62 & 97.32 \\
    Indonesian        & 99.11 & 85.64 & 99.72 & 97.65 \\
    Italian           & 98.25 & 83.62 & 99.89 & 98.28 \\
    Japanese*         & 95.71 & 88.47 & 97.35 & 95.88 \\
    Korean            & 98.36 & 84.40 & 99.36 & 97.10 \\
    Persian           & 96.67 & 88.45 & 95.52 & 93.87 \\
    Polish            & 99.01 & 87.67 & 99.27 & 97.84 \\
    Portuguese        & 96.72 & 88.12 & 90.12 & 95.52 \\
    Romanian          & 99.84 & 77.30 & 99.75 & 98.49 \\
    Russian           & 97.62 & 81.30 & 99.07 & 94.29 \\
    Spanish           & 97.50 & 68.98 & 99.91 & 94.37 \\
    Turkish           & 97.69 & 84.99 & 99.36 & 95.47 \\
    Ukrainian         & 97.46 & 75.89 & 99.67 & 93.70 \\
    Vietnamese        & 96.29 & 79.35 & 99.81 & 91.12 \\
    \noalign{\hrule height 2pt}
    \end{tabular}
    }
    \caption{Case wise accuracies over all languages for each generator : Gemini-1.5-Flash}
    \label{table:112}
\end{table*}
\begin{table*}[!h]
    \centering
    \small{
    \begin{tabular}{ccccc}
    \noalign{\hrule height 2pt}
    \rowcolor{cyan}\textbf{Language }$\downarrow$ & \textbf{Partial cases} & \textbf{Unchanged cases} & \textbf{Rewritten cases} & \textbf{Overall} \\
    \noalign{\hrule height 2pt}
    Arabic            & 96.53 & 92.00 & 99.40 & 95.68 \\
    Chinese*          & 95.05 & 76.30 & 97.63 & 88.07 \\
    Czech             & 96.99 & 78.78 & 94.73 & 94.98 \\
    Dutch             & 94.46 & 78.10 & 91.01 & 94.46 \\
    English           & 96.74 & 90.32 & 99.80 & 95.91 \\
    French            & 97.06 & 74.65 & 97.22 & 94.35 \\
    German            & 96.69 & 76.14 & 98.43 & 94.77 \\
    Greek             & 95.78 & 79.75 & 96.39 & 92.69 \\
    Hebrew            & 95.40 & 83.36 & 97.30 & 93.86 \\
    Hindi             & 96.25 & 92.00 & 99.35 & 95.29 \\
    Indonesian        & 96.67 & 83.15 & 96.32 & 95.55 \\
    Italian           & 97.34 & 82.59 & 99.30 & 96.01 \\
    Japanese*         & 97.05 & 87.46 & 94.90 & 96.12 \\
    Korean            & 96.65 & 82.64 & 97.39 & 95.02 \\
    Persian           & 94.75 & 89.34 & 98.55 & 92.34 \\
    Polish            & 96.69 & 87.13 & 93.69 & 95.36 \\
    Portuguese        & 96.37 & 88.20 & 93.68 & 94.69 \\
    Romanian          & 97.22 & 77.49 & 97.42 & 95.45 \\
    Russian           & 96.64 & 80.58 & 97.90 & 93.09 \\
    Spanish           & 95.54 & 69.92 & 98.82 & 92.76 \\
    Turkish           & 91.99 & 80.61 & 98.47 & 91.99 \\
    Ukrainian         & 96.21 & 74.58 & 97.16 & 93.20 \\
    Vietnamese        & 92.42 & 78.44 & 98.60 & 88.25 \\
    \noalign{\hrule height 2pt}
    \end{tabular}
    }
    \caption{Case wise accuracies over all languages for each generator : Mistral-Large-2411}
    \label{table:113}
\end{table*}
\begin{table*}[!h]
    \centering
    \small{
    \begin{tabular}{ccccc}
    \noalign{\hrule height 2pt}
    \rowcolor{cyan}\textbf{Language }$\downarrow$ & \textbf{Partial cases} & \textbf{Unchanged cases} & \textbf{Rewritten cases} & \textbf{Overall} \\
    \noalign{\hrule height 2pt}
    English           & 97.10 & 91.08 & 99.71 & 96.64 \\
    French            & 95.53 & 72.58 & 99.49 & 93.94 \\
    German            & 94.98 & 77.21 & 99.50 & 94.29 \\
    Portuguese        & 92.66 & 89.06 & 98.17 & 94.03 \\
    Spanish           & 94.79 & 72.31 & 99.80 & 93.99 \\
    \noalign{\hrule height 2pt}
    \end{tabular}
    }
    \caption{Case wise accuracies over all languages for each generator : Perplexity-Sonar-Large}
    \label{table:114}
\end{table*}
\end{document}